\documentclass{article}

\usepackage{arxiv}

\usepackage[utf8]{inputenc} 
\usepackage[T1]{fontenc}    
\usepackage{hyperref}       
\usepackage{url}            
\usepackage{booktabs}       
\usepackage{amsfonts}       
\usepackage{nicefrac}       
\usepackage{microtype}      
\usepackage{lipsum}
\usepackage{graphicx}
\graphicspath{ {./images/} }

\title{Adapting Natural Language Processing Models Across Jurisdictions: A Pilot Study in Canadian Cancer Registries}

\author{
 Jonathan Simkin \\
  School of Population and Public Health\\
  University of British Columbia\\
  Vancouver, BC \\
  \texttt{jonathan.simkin@bccancer.bc.ca} \\
   \And
 Lovedeep Gondara \\
  School of Population and Public Health\\
  University of British Columbia\\
  Vancouver, BC \\
  \texttt{lgondara@mail.ubc.ca} \\
  \And
 Zeeshan Rizvi \\
  Newfoundland \& Labrador Health Services\\
  St. John's, NL \\
  \texttt{Zeeshan.Rizvi@nlhealthservices.ca} \\
    \And
 Gregory Doyle \\
  Newfoundland \& Labrador Health Services\\
  St. John's, NL \\
  \texttt{Gregory.Doyle@nlhealthservices.ca} \\
    \And
Jeff Dowden \\
  Newfoundland \& Labrador Health Services\\
  St. John's, NL \\
  \texttt{Jeff.Dowden@nlhealthservices.ca} \\
    \And
Dan Bond \\
  Newfoundland \& Labrador Health Services\\
  St. John's, NL \\
  \texttt{Dan.Bond@nlhealthservices.ca} \\
    \And
 Desmond Martin \\
  Newfoundland \& Labrador Health Services\\
  St. John's, NL \\
  \texttt{Desmond.Martin@nlhealthservices.ca} \\
    \And
 Raymond Ng \\
  Data Science Institute\\
  University of British Columbia \\
  \texttt{rng@cs.ubc.ca} \\
}

\begin{document}
\maketitle
\begin{abstract}
Population-based cancer registries (PBCRs) depend on pathology reports as their primary diagnostic source, yet manual abstraction is resource-intensive and contributes to delays in cancer surveillance. While transformer-based NLP systems have improved local registry workflows, their ability to generalize across jurisdictions with differing reporting conventions remains poorly understood. We present the first cross-provincial evaluation of adapting BCCRTron, a domain-adapted transformer model developed at the British Columbia Cancer Registry (BCCR), alongside GatorTron, a biomedical transformer model, for cancer surveillance in Canada. Our training dataset consisted of approximately 104,000 and 22,000 de-identified pathology reports from the Newfoundland \& Labrador Cancer Registry (NLCR) for Tier 1 (cancer vs. non-cancer) and Tier 2 (reportable vs. non-reportable) tasks, respectively. Both models were fine-tuned using complementary synoptic-focused and diagnosis-focused report section input pipelines. Across NLCR test sets, the adapted models maintained high performance, demonstrating that transformers pretrained in one jurisdiction can be efficiently localized to another with modest fine-tuning. To further improve sensitivity, we combined the two models using a conservative OR-ensemble. The ensemble achieved Tier 1 recall of 0.99 and reduced missed cancers to 24, compared with 48 and 54 for the standalone models. For Tier 2, the ensemble achieved 0.99 recall and reduced missed reportable cancers to 33, compared with 54 and 46 for the individual models, while maintaining high precision (0.99) and F1 (0.99). These findings demonstrate that an ensemble combining complementary text representations can substantially reduce missed cancers and improve error coverage in cancer-registry NLP. We implement a privacy-preserving workflow in which only model weights, not patient-level data, are shared between provinces, supporting interoperable NLP infrastructure and a future pan-Canadian foundation model for cancer pathology and registry workflows.
\end{abstract}


\section{Introduction}
Cancer is the leading cause of morbidity and mortality worldwide, accounting for nearly 10 million deaths in 2020 alone \cite{sung2021global}. Cancer prevention and control depend on timely, high-quality data from population-based cancer registries, which underpin surveillance, policy and program evaluation, and population oncology research \cite{parkin2006evolution}. Registries collect data from pathology reports, which serve as the primary source record to confirm cancer \cite{blumenthal2020informatics,parkin2006evolution}. However, manual abstraction processes are time-consuming, resource-intensive, and introduce delays in reporting and availability of data that can exceed several months, in many cases, up to 24 months after a diagnosis \cite{santos2022automatic}.

Natural language processing (NLP) and large language models (LLMs) have emerged as viable solutions for automating information extraction from pathology text \cite{santos2022automatic,gondara2024classifying, gondara2025elm}, where Santos et al. confirms that interest in deep learning NLP for cancer pathology is accelerating\cite{santos2022automatic}. The British Columbia Cancer Registry (BCCR) developed deep learning NLP pipelines using domain-adapted transformer models, such as BlueBERT\cite{peng2019transfer} and GatorTron\cite{yang2022gatortron}, which are fine-tuned on local pathology reports. The BCCR achieved over 98 percent recall in classifying reportable tumors, significantly outperforming legacy rule-based systems\cite{gondara2024classifying}. Additionally, the BCCR developed an ensemble of small and large language models for the classification of the anatomical tumor site documented within the pathology reports\cite{gondara2025elm}. 

Whether such models generalize across jurisdictions remains unclear. Each Canadian province use different laboratory information systems and may vary in narrative conventions, although core medical terminology is largely standardized. Smaller or less-resourced registries could benefit greatly from reusing an existing, well-validated model rather than building one from scratch. Prior work demonstrated the promise of transfer learning to adapt pathology-extraction models from one cancer registry to another\cite{alawad2019deep}. Additionally, emerging studies have proposed federated learning as a strategy to train models collaboratively across institutions while preserving data privacy\cite{festag2020privacy}. However, few studies have tested these approaches across provincial cancer registries and in real-world environments.

This study presents a pilot project evaluating the feasibility of adapting BCCR's NLP pipelines for use in the Newfoundland and Labrador Cancer Registry (NLCR). We aimed to replicate BCCR’s two-tier classification architecture for tumour reportability\cite{gondara2024classifying}:
\begin{itemize}
    \item Tier 1 (T1): Classifying pathology reports as “cancer” or “non-cancer”;
    \item Tier 2 (T2): Among cancer reports, identifying those that are “reportable” to the registry.
\end{itemize}

We assess whether a conservative ensemble of two fine-tuned language models, BCCRTron and GatorTron, can accurately perform T1 and T2 classifications on NLCR data.

\section{Methods}
\label{sec:headings}
\subsection{Dataset Description}
Our training dataset consisted of approximately 104,000 and 22,000 de-identified electronic pathology reports from the NLCR for T1 and T2 classification tasks, respectively. These reports were curated, labelled, and prepared by NLCR Oncology Data Specialists. All records were de-identified prior to processing and analysis. Each report included multiple sections, such as “synoptic,” “diagnosis,” and “specimen,” with variable structure and content across hospitals and time periods. The reports were randomly sampled from 2022 and 2023 diagnosis years. For T1, the dataset consisted of 21\% cancer and 79\% non-cancer. For T2, the dataset consisted of 80\% reportable and 20\% non reportable.

\subsection{Training data}
To prioritize higher recall for cancer and reportables cases, goals aligned with cancer registry objectives, and to address class imbalance, we constructed training datasets using random undersampling with out replacement. For T1, non cancers cases were undersampled using the following formula: 
\begin{equation}
N(non cancer) = N(cancer)*0.8 
\end{equation}
For T2, non-reportable case were undersampled using: 
\begin{equation}
N(reportable) = N(non reportable)*1.2. 
\end{equation}

This approach resulted in training dataset sizes of n=31,694 for T1 and n=7,821 for T2. 

\subsection{Training data}
To ensure robust evaluation, we used multiple test datasets spanning different time periods for both T1 and T2 models. The initial test datasets were created using an 80-20 training/test split with the full datasets (before undersampling), leading to n=20,777 for T1 and n=4,409 for T2. This ensured that the initial test sets reflected the true, real-world data distribution. Additional test sets, that were not part of the original datasets were used for further evaluation, these test sets were sampled from the diagnosis year of 2024 with n=18,549 for T1 and n=3,078 for T2. The cancer/no cancer distribution was 17\%/83\%  in the additional test set for T1 and 100\%/0\% reportables/non-reportables in the additional test set for T2. 

\subsection{Models and Fine-Tuning}
We employed two pretrained transformer-based language models:
\begin{itemize}
    \item GatorTron\cite{yang2022gatortron}, a large biomedical language model trained on extensive clinical and electronic health record corpora;
    \item BCCRTron\cite{gondara2024classifying,gondara2025smalllarge}, Gatortron\cite{yang2022gatortron} further pre-trained on BC Cancer Registry pathology text using $\sim$1.1M pathology reports.
\end{itemize}

Each model was fine-tuned on the NLCR data using the outputs of each preprocessing pipeline separately (detailed below), yielding four fine-tuned models in total: two per tier (T1 and T2, based on two input pipelines). Hyperparameters were selected following established best practices to reduce over fitting, including limiting training epochs\cite{devlin2019bert}; full details are provide in the Supplement (Table S1). Fine-tuning was performed on a desktop machine, with Intel Xeon CPU, 64 GB RAM, and a single NVIDIA RTX 4090 GPU (24 GB VRAM).  We refer to the finetuned models with their original names (i.e. BCCRTron and Gatortron).

 Additionally, we used different text preprocessing methods to ensure model output diversity for an efficient ensemble, which we detail next.
 
\subsubsection{Preprocessing and input pipelines}
The text was normalized by lowercasing, removing punctuation, and tokenizing using model-specific tokenizers. To support an ensemble strategy, we developed two distinct preprocessing pipelines that introduced variation in how each model processed the same pathology report. The aim was to train models that would emphasize different aspects of the input, improving robustness when used in combination.

\begin{itemize}
    \item Pipeline A extracted and prioritized the structured synoptic section, which typically contains semi-structured, checklist-style data aligned with standardized cancer reporting formats. This pipeline was used as an input for BCCRTron.
    \item Pipeline B extracted and prioritized the diagnosis section, consisting of free-text histopathological descriptions and clinical impressions. This pipeline was used as an input for Gatortron.
\end{itemize}
This dual-pipeline approach ensured that each model would be exposed to complementary information during fine-tuning. The rationale was that if a report truly belonged to the negative class (e.g., non-cancer in T1 or non-reportable in T2), both models would independently assign a negative label, despite being trained on different views of the report. In contrast, if only one model detected relevant signal, the ensemble could still flag the report as positive, aligning with our goal of maximizing sensitivity.

\subsubsection{Classification Tasks and Ensemble Strategy}
We replicated the same architecture and modeling approach for both classification tiers:
\begin{itemize}
    \item T1: Classify each report as “cancer” or “non-cancer”;
    \item T2: Among reports classified as cancer, classify whether the report represents a “reportable” cancer to the registry. 
\end{itemize}
	
For both tiers, we adopted an ensemble strategy with BCCRTron and GatorTron. A report was labeled as positive (cancer or reportable, depending on tier) if at least one of the two models in the given pipeline predicted a positive class. The “OR” ensemble strategy, proposed by Gondara et al. (5), was chosen to prioritize sensitivity and minimize false negatives, consistent with registry goals.

\subsubsection{Evaluation}
We evaluated model performance using hold-out and additional test sets. For T1, gold standard cancer vs. non-cancer labels were based on manual abstraction by trained NLCR staff, which serves as the operational gold standard. For T2, ground truth labels were drawn from a subset of cancer-positive reports manually annotated as reportable or not. We report standard performance metrics: accuracy, sensitivity, specificity, precision, and F1 scores (micro and macro) for both tiers. 

\section{Results}
\subsubsection{T1: Cancer vs. Non-Cancer Classification}
All models achieved high performance, indistinguishable for recall, our priority, using two decimal places. For the cancer labels, the ensemble model achieved recall of [99\%], precision of [88\%], and F1 score of [93\%]. Detailed results are shown in Table 1. 

After fine-tuning on NLCR data, BCCRTron achieved [99\%] recall and [91\%] precision. The Gatortron model (GatorTron fine-tuned on NLCR data) yielded [99\%] recall and [89\%] precision.

Looking at absolute number of missed cancers, we see that the ensemble model performs the best, missing half the number of cases compared to other models, where the combined model missed 24 cases, compared to 48 and 54 missed by Gatortron and BCCRTron respectively. Note that similar performance was observed on additional test sets.

\begin{table}[htbp]
\centering
\caption{T1 performance metrics for individual models and the ensemble.}
\begin{tabular}{lccc}
\toprule
\textbf{Model (class)} & \textbf{Recall} & \textbf{Precision} & \textbf{F1 score} \\
\midrule
GatorTron (non cancer)  & 0.97 & 1.00 & 0.98 \\
GatorTron (cancer)      & 0.99 & 0.89 & 0.94 \\
BCCRTron (non cancer)   & 0.97 & 1.00 & 0.99 \\
BCCRTron (cancer)       & 0.99 & 0.91 & 0.95 \\
Combined (non cancer)   & 0.96 & 1.00 & 0.98 \\
Combined (cancer)       & 0.99 & 0.88 & 0.93 \\
\bottomrule
\end{tabular}
\label{tab:t1_metrics}
\end{table}

\subsubsection{T2: Reportable vs. Non-Reportable Classification}
For T2 classification, the ensemble achieved [99\%] recall, [99\%] precision, and an F1 score of [99\%] on the cancer-positive initial NLCR test set for the reportable class.  Whereas the finetuned BCCRTron and Gatortron achieved the recall, precision, and F1 scores of [99\%,100\%,99\%] and [98\%,100\%,99\%] respectively. Detailed results are shown in Table 2.

As with T1, the ensemble achieved the highest recall, while balancing precision, than any individual model. Where the combined model missed 33 reportables compared to 54 and 46 by Gatortron and BCCRTron respectively. This again reflects the intended bias toward recall in registry workflows. Note that similar performance was observed on additional test sets.

\begin{table}[htbp]
\centering
\caption{T2 performance metrics for individual models and the ensemble.}
\begin{tabular}{lccc}
\toprule
\textbf{Model (class)} & \textbf{Recall} & \textbf{Precision} & \textbf{F1 score} \\
\midrule
GatorTron (non reportable)   & 0.99 & 0.94 & 0.96 \\
GatorTron (reportable)       & 0.98 & 1.00 & 0.99 \\
BCCRTron (non reportable)    & 0.99 & 0.95 & 0.97 \\
BCCRTron (reportable)        & 0.99 & 1.00 & 0.99 \\
Combined (non reportable)    & 0.98 & 0.96 & 0.97 \\
Combined (reportable)        & 0.99 & 0.99 & 0.99 \\
\bottomrule
\end{tabular}
\label{tab:t2_metrics}
\end{table}

\section{Discussion}
This study demonstrates the feasibility of sharing and adapting NLP models across provincial cancer registries in Canada. To our knowledge, this is the first instance of interprovincial collaboration involving the direct reuse and local adaptation of domain-specific language models for cancer surveillance. By deploying a plug-and-play model architecture developed at BCCR and fine-tuning it on NLCR data, we achieved strong performance in both T1 (cancer vs. non-cancer) and T2 (reportable vs. non-reportable) classification tasks. The ensemble model, built from two domain-adapted transformers with complementary input pipelines, outperformed individual models in sensitivity while maintained acceptable specificity. This performance and our approach suggests a practical, inexpensive, and robust path to scalable NLP deployment across cancer registries.

Our findings build on a growing body of work demonstrating the value of transfer learning in clinical NLP, particularly for pathology information extraction. Alawad et al. evaluated several cross-registry transfer strategies and found that fine-tuning a shared model on a second registry’s data substantially improved extraction performance, particularly for rare entities. This was achieved within a distributed, privacy-preserving setup and approached the accuracy of centrally pooled models\cite{alawad2019deep}. In our study, we observed that BCCRTron, pre-trained on BC cancer pathology text, performed well when finetuned and applied to NL data. This supports the broader idea that transformer models trained in one institutional context can be effectively localized to another with finetuning, enabling efficient reuse of high-performing infrastructure across jurisdictions.

A second key design of our study was the use of an ensemble architecture combining two language models each trained on a distinct textual input. BCCRTron prioritized synoptic sections, while GatorTron focused on diagnostic text, providing different views of the same report. This design increased representational diversity and allowed the ensemble to capture cases that may have been missed by individual models. Consistent with registry priorities, we adopted an “OR” logic to maximize sensitivity, even at a modest cost to specificity. This mirrors prior findings by Gondara et al., who achieved over 98\% sensitivity in tumour reportability classification using an ensemble of BlueBERT and GatorTron applied to BC pathology reports\cite{gondara2024classifying}. Other studies have demonstrated that ensemble methods can improve robustness, performance, and generalizability in clinical NLP tasks\cite{li2024ensemble,naderalvojoud2024improving,yang2023oneLLM}. However, ensemble methods also introduce important limitations, including increased computational cost and reduced interpretability\cite{ganaie2022ensemble,mohammed2023comprehensive}. Moreover, ensemble methods are most effective when constituent models exhibit sufficient diversity, supporting our approach to train models on distinct textual views to maximize complementary strengths and overall robustness\cite{naderalvojoud2024improving}.

Preserving data privacy and respecting jurisdictional data boundaries are core challenges in healthcare. In our study, all pathology reports were fully de-identified prior to training, and model fine-tuning occurred entirely within provincial boundaries. Only model weights were shared and updated, enabling collaboration without cross-jurisdictional data transfer. This reflects a pragmatic, privacy-preserving alternative to centralized training, aligned with distributed learning settings\cite{alawad2019deep,zerka2020systematic}. Federated learning (FL) is frequently proposed as a formalized alternative to collaborative model development without data sharing across institutions\cite{peng2024federated,zhang2024recent}. While promising, the practical implementation of FL in clinical NLP faces several challenges such as coordinating and managing real-time training across sites, aligning model architectures, and mitigating communication overhead\cite{li2025from,zhang2024recent} Our approach, sharing model weights but training entirely within provincial boundaries, offers a simpler, pragmatic pathway aligned with these goals. Active research continues to explore methods at reducing the practical burden of FL\cite{li2025from,li2025selective}. 

\subsection{Limitations}
This study has several limitations. First, it focused on adapting models to a single target registry, and further validation across diverse jurisdictions is needed to assess generalizability. Second, while ensemble methods improved sensitivity, they introduced added computational complexity, which may affect deployment in resource-constrained settings. Addressing these limitations is an important focus for future work as we explore broader implementation.

\subsection{Future directions}
This study highlights a viable path toward collaborative NLP for Canadian cancer registries, and in the healthcare domain in general. While our study focused on adapting existing models to a single new jurisdiction, the same framework could be extended to support a pan-Canadian foundation model co-developed through distributed or federated approaches. Realizing this at scale will require not only model development but also shared data standards, consistent evaluation protocols, and formal agreements around model sharing and privacy protection. Moreover, integrating these models into operational workflows will require alignment with registry information systems and ongoing validation by subject matter experts. Importantly, this work aligns with the goals of the Pan-Canadian Cancer Data Strategy\cite{pan_canadian_cancer_data_strategy}, which calls for innovation, interoperability, and timely cancer data to support population health. Nevertheless, our findings suggest that cross-jurisdictional reuse of language models is not only technically feasible but also strategically aligned with national efforts to modernize health data infrastructure and improve the timeliness of cancer surveillance.

\subsection{Conclusion}
In conclusion, this study offers practical evidence that pretrained NLP models can be adapted across jurisdictions to support timely and automated cancer surveillance. By enabling reuse of high-performing infrastructure while preserving data locality, our approach provides a feasible and scalable pathway for modernizing registry workflows in Canada and beyond.

\subsection{Acknowledgements}
We thank Jennifer Gushue and team at the Newfoundland \& Labrador Cancer Registry for their expertise, data preparation support, and essential contributions to this work. We also thank subject matter experts and registry leadership at the BC Cancer Registry for their support. This study was supported by the Canadian Partnership Against Cancer through the Pan-Canadian Cancer Data Strategy, whose funding enabled cross-jurisdictional collaboration and methodological advancement in cancer registry modernization.

\bibliographystyle{unsrt}  
\bibliography{references}  



\end{document}